# Computerization of Clinical Pathways: A Literature Review and Directions for Future Research

Ayman Alahmar[*,1] and Ola Alkhatib[2]

[*]Corresponding Author, aalahmar@lakeheadu.ca, ORCID: 0000-0003-4011-1023
[1]*Department of Software Engineering, Lakehead University, Thunder Bay, Ontario, Canada*
[2]*Department of Computer Science, Lakehead University, Thunder Bay, Ontario, Canada*

**Abstract** Clinical Pathways (CP) are medical management plans developed to standardize patient treatment activities, optimize resource usage, reduce expenses, and improve the quality of healthcare services. Most CPs currently in use are paper-based documents (i.e., not computerized). CP computerization has been an active research topic since the inception of CP use in hospitals. This literature review research aims to examine studies that focused on CP computerization and offers recommendations for future research in this important research area. Some critical research suggestions include centralizing computerized CPs in Healthcare Information Systems (HIS), CP term standardization using international medical terminology systems, developing a global CP-specific digital coding system, creating a unified CP meta-ontology, developing independent Clinical Pathway Management Systems (CPMS), and supporting CPMSs with machine learning sub-systems.

*Keywords:* Clinical pathway, Computerization, HL7, Literature review, Machine Learning, Modeling, Ontology, Semantic web, SNOMED CT, Standardization.

## INTRODUCTION

CLINICAL Pathways (CP) first emerged in hospitals in the USA in the 1980s when Karen Zander, Kathleen Bower, and Mary Etheredge coined the term at the New England Medical Center in Boston [1]. The concept of CPs has its roots in management theories that focus on improving the quality of processes in business, such as Business Process Reengineering (BPR), Program Evaluation and Review Technique (PERT), and Critical Path Method (CPM). Therefore, the CP concept was an initiative to adopt successful management concepts in healthcare [2]. Following their initial launch in the USA, CPs were used in the UK, after which the concept was adopted internationally. CP is defined slightly differently in various research articles. Table I presents some CP definitions.

TABLE I. CP DEFINITIONS IN THE LITERATURE

| CP Definition | Reference |
|---|---|
| Comprehensive methods of planning, delivering, and monitoring different healthcare services provided to patients. | [3] |
| Methodology for the mutual decision making and organization of care for a well-defined group of patients during a well-defined period. | [4, 5] |
| Optimal sequencing and timing of interventions by physicians, nurses, and other staff for a particular diagnosis or procedure, | [6, 7] |

| designed to minimize delays and resource utilization and to maximize the quality of care. | |
|---|---|

CPs are becoming popular in hospitals around the world due to their extensive benefits [8-11]. CP benefits include: reducing patients' hospitalization cost and Length of Stay (LOS), improving patient satisfaction and outcomes, minimizing treatment variations, optimizing resource utilization, increasing patient participation in medical procedures, and improving communication between healthcare professionals [8-11].

CPs are also considered an essential tool for ensuring that the most recent evidence in clinical guidelines is used [12].

It is estimated that healthcare organizations could reduce costs by 30% [13] or by 30-50% [14] if they adopt the best IT and quality management practices that discontinue the overuse of resources and reduce waste. To achieve these financial benefits, the proper application of CPs is essential since CP systems are at the core of best management practices in hospitals.

Despite the increasing popularity of CPs, they are still used mainly in paper-based formats. CP computerization brings great advantages because many benefits of CPs can be fully realized through software system automation. Due to the importance of CP computerization, the objective of this work is to present a literature review on CP computerization followed by suggestions for future research. Using popular research digital libraries (such as Science Direct, PubMed, IEEE Xplore, and Google Scholar), we searched phrases and terms such as clinical pathway, care pathway, computerization, automation, modeling, ontology, etc. in order to retrieve existing studies (up to mid-2020) on this important topic.

Our literature review on CP computerization is presented below. The selected studies were categorized by adopting both semantic-based and non-semantic-based methods, followed by suggestions for future research and a conclusion section.

## CP COMPUTERIZATION

The literature review reveals that several studies addressed the computerization of CPs. Some research was based on non-semantic web (i.e., traditional) information systems, while most recent research articles adopted a semantic web approach in which CPs were modeled using ontology engineering methods. Semantic modeling is a relatively recent approach for knowledge representation in software engineering and is widely used for data management in health informatics [15]. Web Ontology Language (OWL) and Semantic Web Rule Language (SWRL) are central components of semantic models [16, 17]. OWL and SWRL help in using semantic web statements and rules to define classes, relationships, and domain constraints for the purpose of domain modeling (e.g., an ontology to model lung cancer) [18]. The literature review revealed that various diseases and medical conditions were considered in CP modeling studies. Since semantic-based studies share similar modeling techniques, we begin by addressing selected articles among them, followed by discussing selected studies that applied the non-semantic, traditional CP modeling approach.

### A. *Semantic-based Methods*

Tehrani et al. [19, 20] pointed out that the development of CPs in situations where processes are complex must combine organizational semiotics and ontology-based modeling.

Organizational semiotics treats organizations as information systems in which information is created, processed, distributed, stored, and used [21]. In their method, they interviewed medical staff members and used semantic analysis to develop an ontology that represents the semantics of the CP concepts, their relationships and behavior patterns of staff members and physicians. Next, they used the norm analysis method to extract and analyze patterns of healthcare activities and informal safety norms that affect patient safety and CP outcome. Norms in semiotic approaches specify the possible behavior patterns. For instance, the nurse is "permitted," "obliged," or "prohibited" to execute an action (called deontic operators). Norms are described formally using the format:

**Whenever** (condition) **If** (state) **Then** (agent) **Is** (deontic operator) **To** (action)

For example, Norm N1 can be defined as:

**Whenever** (the patient is assessed for venous thromboembolism)
**If** (there is bleeding risk)
**Then** (doctor) **Is** (permitted) **To** (give prophylaxis)

The authors suggested that "generating a CP ontology that is enhanced by formal patterns of human behavior and rules that govern the actions identified in the ontology" reduces human errors associated with complex situations that require human decision and patient-specific customization. The ontology can then be integrated with an electronic medical record (EMR) system.

Fudholi et al. [22] addressed an ontology to model CPs that consists of the major classes: CP, person, clinical category, record, and organizational structure. The model was proposed to verify if the treatment processes complied with the requirements of the CP. For instance, by querying both the patient's recorded data and the CP ontology, the model checks if the recorded medical data comply with the steps/interventions specified in the CP. As an evaluation of their model, the authors developed typhoid fever and dengue fever CP ontologies and queried them through an ontology query language to perform certain compliance checks. Their results revealed that they have not considered standardized CPs in their model. For instance, they referred to the CP lab tests in non-standardized terms like HT and HB, which indicates that a local terminology approach was used in the developed ontology.

Wang et al. [23] proposed a method that supports EMR with a CP system through SNOMED CT linking between terms that are equivalent in both systems. Thus, the CP system can be integrated programmatically with different EMRs. The programmers need to generate Resource Description Framework (RDF) statements from the EMR database and then use an ontology editing tool to write the statement of relations between EMR terms and CP terms. A limitation of their work is the tedious programming work that is required to link the CP with the EMR. In addition, modifications to the EMR dictionary would cause the EMR-CP integration to be lost, which would necessitate re-programming.

Liu et al. [24] presented a semantic-based method (based on ontology) for CP monitoring. Their goal was to establish communication between the EMR and CP with the capability of monitoring CP execution and display reminders to physicians/nurses regarding CP activities.

The CP used in their system was for unstable angina obtained from the cardiology department of a hospital in China. Although they successfully built a CP component that feeds data/reminders to EMR, the system was not performing as a fully-independent CP system.

Abidi et al. [25, 26] proposed an ontology-based computerization model for prostate cancer, and studied the merging of prostate cancer CPs from various hospitals in Canada. In their model, they represented hospital-specific CPs using an ontological model, and aligned the common activities between CPs from different hospitals to obtain a common CP. The resulting model merges common healthcare activities while permitting to have unique hospital-specific activities. Their model is useful for performance analysis; however, their method is effective only for small-scale unification and is not practical for a larger scale multi-hospital approach.

Daniyal et al. [27] followed a similar approach to develop an ontology-based prostate cancer CP that integrates different CPs to create a common CP for prostate cancer. The authors also integrated the developed CP in a computerized system for prostate cancer that automated a combined flowchart for three different CPs.

Hu et al. [28] recommended an ontological approach in which CPs were modeled using ontologies and CP rules were modeled using SWRL. Using this method, the CP system could reason over the information collected and rules. The authors used a CP meta (or general) ontology that defines common concepts required in disease-specific CPs. To test their model, they developed a CP for lobectomia pulmonalis and integrated it with an EMR system called IZANAMI. An illustration of the structure of their system is shown in Figure 1.

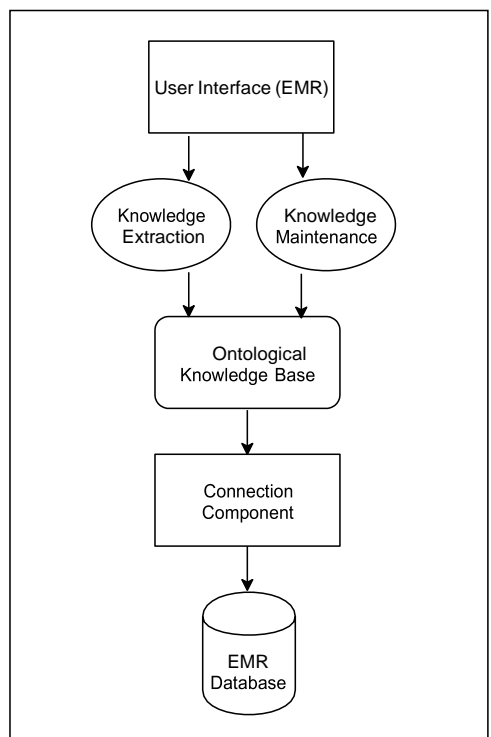

**Fig. 1.** The structure of the system developed by Hu et al. [28].

The model was successful in presenting CP steps to healthcare staff members. The limitation of their approach is that their meta-ontology was a hospital-specific ontology since its modeling was based on the terms available in local CPs without standardization of CP terminology using international terminology. In addition, their CP system was totally embedded inside the EMR system, as shown in the structure of their system.

Alexandrou et al. [29] addressed a CP ontology model that was comprised of three parts in one ontology: the CP part, the business part, and the quality assurance part. The authors applied their model on human papillomavirus patients; however, standardization of CP terms was not considered in their study.

Ye et al. [30] proposed a CP ontology model in which they modeled time intervals between CP tasks using the entry sub-ontology of time. In their implementation, they used a CP for cesarean section from a hospital in Shanghai, China.

Hu et al. [31] modeled CPs based on an ontology schema that consisted of four main units: category of care, timeline, outcome criteria, and variance record. The authors demonstrated that the ontology-based method is suitable to model CPs by comparing the CP concepts with ontology. For instance, both CP and ontology are formalization methods (i.e., CP formalizes clinical care processes in healthcare and ontology formalizes concepts in a domain). Their work was limited to ontology-based CP modeling and deemed it as a successful modeling approach. No system was built to use the ontology in this study.

Alahmar et al. [32-34] developed a CP automation framework that centralizes CPs in healthcare by positioning CP management systems at the center of health information systems and integrating them with existing systems through Health Level 7 (HL7) standard. Thus, the framework can serve as a platform for health informatics by capturing all CP data. In addition, the authors developed an international CP identification code based on SNOMED CT terminology system (i.e., digital coding system for CPs). The digital coding system facilitates CP-based decision support in hospitals and improves CP data collection, sharing, auditing, and quality management. The authors modeled CPs based on a higher-level CP ontology (meta-ontology) that models generic CP knowledge using standardized vocabulary to support semantic interoperability (see Figure 2). Disease-specific CPs can be independently extended and specialized from the higher-level ontology. The digital coding system facilitates CP-based decision support in healthcare and ensures the independence of CP management systems by including a data repository and decision-support/data analytics component.

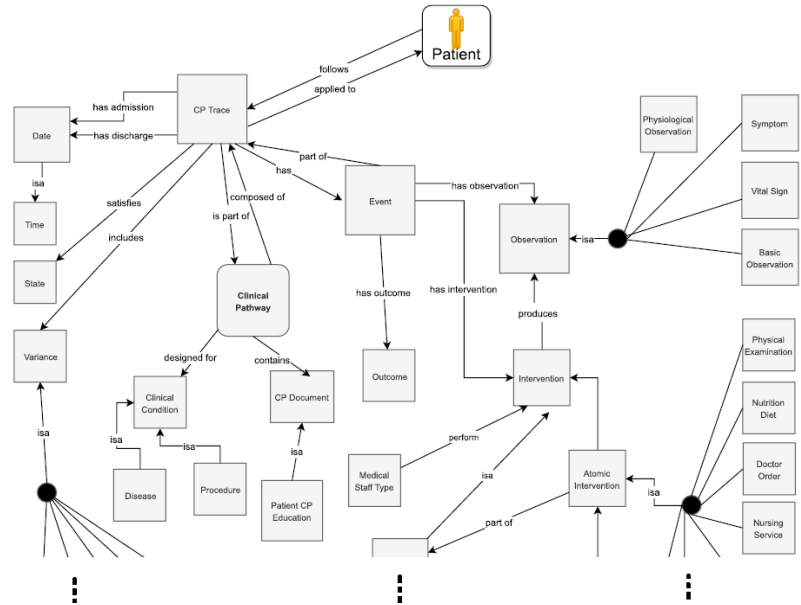

**Fig. 2.** Part of the CP meta-ontology (higher-level ontology) [33].

Ma et al. [35] proposed a modeling method to CP computerization based on process ontology. Their method was based on developing an ontology model of the diagnosis and treatment processes. In the ontology, the CP instance was transformed into the linear combination of the ontology for the task of diagnosis, while this ontology was extracted from clinical doctor's recommendation. The system requires users to describe the action tasks and check items in doctor's advice to build a computerized CP. This method reduces the cost, time and staff required for the development of the CP instances. Figure 3 depicts the framework of the developed CP information system in this work.

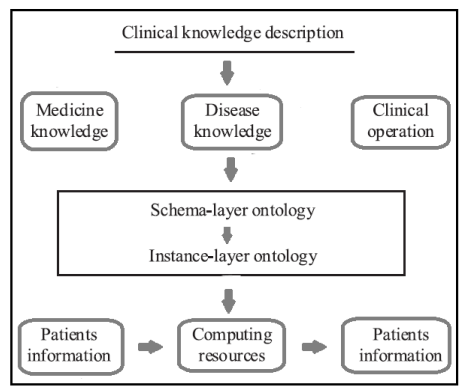

**Fig. 3.** Framework of CP information system [35].

### B. Non-semantic-based Methods

Below are studies that we reviewed that followed the non-semantic modeling approach. Studies in this category modeled CP mainly by programming additional chosen CP data fields into the EMR systems in order to computerize CP steps.

Hoelscher et al. [36] modeled a computerized CP for infectious diseases within an EMR

system. Their goal was to implement an improved rapid-deployment decision support strategy for the detection and treatment of emerging and re-emerging infectious diseases. Using the Plan-Do-Study-Act (PDSA) rapid cycle improvement model, the computerization process was implemented and monitored.

Smulowitz et al. [37] developed a clinical decision support tool within a system for emergency department. Their objective was limited to flagging patients who were required to follow a CP for chest pain, which was labelled as "HEART pathway."

Gibbs et al. [38] developed a framework for building an online CP that can be used directly by patients. They applied their framework on a CP for chlamydia infection. However, their approach was limited in scope to the considered disease.

Blaser et al. [39] developed a prototype system by using an embedded tool within Orbis/OpenMed-system (an EMR system used in Germany). Although they managed to add selected CP data into the EMR, the CP functionality performed well only within the EMR and could not be utilized as an independent CP management system.

Bernstein et al. [40] pointed out that CPs are not well integrated with EMRs. Therefore, they proposed a method that made the patient's position in the CP visible in such a way so that each CP would have a SNOMED CT link to the EMR system. Their SNOMED CT linking was limited to a top-level integration between the EMR and major steps of the CP. For example, the 'laboratory tests' stage in the CP was considered to be a single node linked to the EMR to show that the patient has reached this stage without considering the detailed CP contents. Such linking might help determine the patient's position in the CP; however, it cannot help to capture all detailed CP data to reduce missing data and improve the data mining outcome.

Katzan et al. [41] developed a stroke CP program that was integrated within an Epic EMR (a commercial EMR developed by Epic Systems Corporation) [42]. Epic programming contractors were involved in this work to develop the program and to customize the Epic EMR screens to include CP-specific steps. The modified interface saved time for clinicians by reducing unnecessary data entry based on the CP. Integrating the CP within the EMR reminded healthcare providers of certain CP guidelines that might have been forgotten, which reduced possible human errors. They reported that the integration of CP with EMR was successful overall; however, not all data fields were captured, and not all features functioned as planned. For example, an anatomic diagram for stroke location did not function within the program despite extensive efforts by the programmers to make it work. In addition, because some of the discharge checklist items did not auto-populate correctly, healthcare providers were not using them [41]. Furthermore, the system was programmatically integrated within the EMR and therefore, it was not an independent electronic CP system.

### SUGGESTIONS FOR FUTURE RESEARCH

In order to offer recommendations for future research, we must analyze the existing literature. A critical analysis of the literature review reveals research gaps, as outlined below.

As described in the literature review, most CP computerization studies consider the CP system as a secondary component in Health Information Systems (HIS). This is because the final target of the computerization process in most articles was the EMR (as the central component) and finding ways to support EMR with CPs. Few recent studies consider the CP as a centralized component in HISs [32-34]. Computerized CPs deserve to be positioned at the center of HISs because CP interventions on patients contain all the data that feed the databases of other information systems (refer to Figure 4). Therefore, more research that focuses on centralizing computerized CPs in HISs is required.

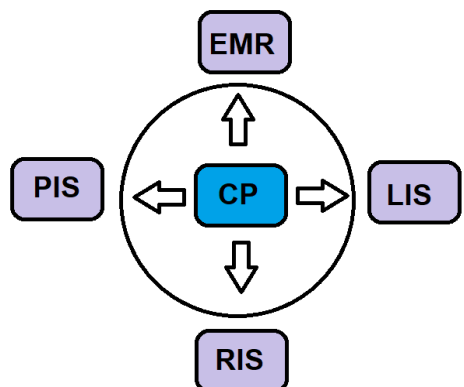

**Fig. 4.** Centralizing computerized CPs in healthcare information systems (IS: Information System, LIS: Laboratory IS, PIS: Pharmacy IS, RIS: Radiology IS, CP: Clinical Pathway and EMR: Electronic Medical Record) [32-34].

The view that CP systems were considered secondary HIS components has resulted in partially-standardized CPs or incomplete digitization of CPs, which are major limitations of the research found in the literature. CPs developed in hospitals contain many local, non-standardized terms. Most studies developed computerized CPs with either a complete lack of CP term standardization, or those that contained only partial standardization. Few recent articles considered full CP term standardization and digitization [32-34]; however, the standardization in these recent studies considered only the SNOMED CT terminology system as the base for CP standardization. Thus, one suggestion for future research is to conduct more studies on CP standardization that consider other terminology systems in the standardization process (e.g., ICD-10, ICD-11 and LOINC). CP standardization improves CP data communication between medical professionals, which reduces human errors in healthcare institutions. Table II below outlines the limitations and advantages of CP computerization without and with full CP term standardization, respectively [42].

**TABLE II.** LIMITATIONS OF CP COMPUTERIZATION WITHOUT FULL CP TERM STANDARDIZATION [42]

| | Computerization without CP Standardization | Computerization with CP Standardization |
|---|---|---|
| CP Integration with HISs | Difficult to integrate due to mismatch between terms on CPs and terms used in HISs. | Easy to integrate since CPs share the same terminology standards with various HISs. |
| CP Update in EMRs | Connection is lost and re-work is required to connect CP to EMR whenever the CP is updated. | Updated standardized terms integrate smoothly with counterpart EMR terms; re-work is not required, and the connection is not lost. |
| CP Sharing Across Different Healthcare Institutions | The terminology conflict creates ambiguity and makes CP sharing impossible. | CP standardization streamlines CP sharing without ambiguities. |

| CP Data Management and Reporting | Complex data management; reports are susceptible to errors due to conflicts. | Simple data management; standardized reports can be generated easily and accurately. |
|---|---|---|

Aside from CP standardization, the present gap between CPs and their integration with HISs can be linked to the fact that CPs do not have an international digital system to identify them and their contents in digital systems. While recent research has developed a digital coding system for CPs and their data [32], this must be an international research effort. Therefore, an important future research recommendation is the collaboration between researchers to advance the CP digital coding system proposed in [32] in order to improve it and facilitate its international adoption. Furthermore, a recognized CP coding system would help integrate computerized CPs with HISs using international integration standards like HL7 (actually, CPMS/HL7 integration is also an important suggestion for future research).

Another research gap is the need for a unified CP meta-ontology. Even though various researchers have proposed the use of a CP meta ontology to model common CP activities, the reality in literature is that there are several meta-ontologies that differ from each other. Therefore, a future research could encompass collaboration between researchers and international health organizations towards developing and maintaining a unified CP meta ontology that can be used in HISs and CP Management Systems (CPMS), worldwide.

Although we could offer many other ideas for future research, but due to space limitation, we submit one final important recommendation that could offer many advantages (last but certainly not least): research targeting the development of advanced and independent CPMSs. In the era of data science and machine learning, independent CPMSs can have an important function beyond the computerization of CPs and integrating them with other HISs. For example, independent CPMSs can be supported with data analytics/machine learning components that facilitate big data mining in healthcare [32-34].

## CONCLUSION

CP computerization has been an active research area since the launch of CPs in hospitals. This research presents literature review on CP computerization and integration with HISs. Although the literature review was not comprehensive, it shed light on the research efforts in this area and highlighted some of the research gaps. The study recommends many useful future research directions, including centralizing CPMSs in HISs, full CP standardization, considering terminology systems other than SNOMED CT in CP standardization, developing an international CP-specific coding system, CPMS/HL7 integration, creating a unified CP meta-ontology, building independent CPMSs, and enhancing CPMSs with data analytics/machine learning components.

### REFERENCES

[1] K. Zander, K. A. Bower, and M. Etheredge, "Nursing case management: blueprints for transformation," Boston: New England Medical Center Hospitals, pp. 1–128, 1987.

[2] R. S. Russell and B. W. Taylor, Operations and Supply Chain Management, 9th Edition. John Wiley & Sons, 2017

[3] M. Khalifa and O. Alswailem, "Clinical pathways: Identifying development, implementation and evaluation challenges," in ICIMTH, pp. 131–134, 2015.

[4]  European Pathway Association, http://e-p-a.org. Accessed: 2022-01-15.

[5]  L. De Bleser, R. Depreitere, K. D. WAELE, K. Vanhaecht, J. Vlayen, and W. Sermeus, "Defining pathways," Journal of nursing management, vol. 14, no. 7, pp. 553– 563, 2006.

[6]  X. Wang, S. Su, H. Jiang, J. Wang, X. Li, and M. Liu, "Short-and long-term effects of clinical pathway on the quality of surgical non-small cell lung cancer care in china: an interrupted time series study," International Journal for Quality in Health Care, vol. 30, no. 4, pp. 276–282, 2018.

[7]  A. Alahmar, E. Mohammed and R. Benlamri, "Application of data mining techniques to predict the length of stay of hospitalized patients with diabetes," IEEE 4ᵗʰ International Conference on Big Data Innovations and Applications (Innovate-Data), pp. 38-43, 2018. DOI: 10.1109/Innovate-Data.2018.00013.

[8]  B. J. Gebhardt, J. Thomas, Z. D. Horne, C. E. Champ, G. M. Ahrendt, E. Diego, D. E. Heron, and S. Beriwal, "Standardization of nodal radiation therapy through changes to a breast cancer clinical pathway throughout a large, integrated cancer center network," Practical radiation oncology, vol. 8, no. 1, pp. 4–12, 2018

[9]  A. K. Lawal, T. Rotter, L. Kinsman, A. Machotta, U. Ronellenfitsch, S. D. Scott, D. Goodridge, C. Plishka, and G. Groot, "What is a clinical pathway? refinement of an operational definition to identify clinical pathway studies for a Cochrane systematic review," BMC medicine, vol. 14, no. 1, p. 35, 2016.

[10]  S. Preston, S. Markar, C. Baker, Y. Soon, S. Singh, and D. Low, "Impact of a multidisciplinary standardized clinical pathway on perioperative outcomes in patients with oesophageal cancer," British journal of surgery, vol. 100, no. 1, pp. 105–112, 2013.

[11]  P. A. Van Dam, G. Verheyden, A. Sugihara, X. B. Trinh, H. Van Der Mussele, H. Wuyts, L. Verkinderen, J. Hauspy, P. Vermeulen, and L. Dirix, "A dynamic clinical pathway for the treatment of patients with early breast cancer is a tool for better cancer care: implementation and prospective analysis between 2002–2010," World journal of surgical oncology, vol. 11, no. 1, p. 70, 2013.

[12]  P. Bjurling-Sj¨oberg, Clinical Pathway Implementation and Teamwork in Swedish Intensive Care: Challenges in Evidence-Based Practice and Interprofessional Collaboration. PhD thesis, Acta Universitatis Upsaliensis, 2018

[13]  W. G. Carnett, "Clinical practice guidelines: a tool to improve care," Journal of nursing care quality, vol. 16, no. 3, pp. 60–70, 2002.

[14]  "Medicare and Medicaid EHR Incentive Programs." http://fusionppt.com. Accessed: 2021-06-18

[15]  C. J. Baker and e. Kei-Hoi Cheung, Semantic web: Revolutionizing knowledge discovery in the life sciences. Reading, Massachusetts: Springer Science and Business Media, 2007.

[16]  "W3C Web Ontology Language (OWL)." https://www.w3.org. Accessed: 2021-10- 30.

[17]  "SWRL: A Semantic Web Rule Language." https://www.w3.org/Submission/ SWRL. Accessed: 2021-10-30.

[18]  J. Hebeler, M. Fisher, R. Blace, and A. Perez-Lopez, Semantic web programming. John Wiley & Sons, 2011.

[19]  J. Tehrani, "Computerisation of clinical pathways," in Healthcare Ethics and Training: Concepts, Methodologies, Tools, and Applications, pp. 1050–1074, IGI Global, 2017.

[20] J. Tehrani, "Computerisation of clinical pathways: Based on a semiotically inspired methodology," in E-Health and Telemedicine: Concepts, Methodologies, Tools, and Applications, pp. 25–48, IGI Global, 2016.

[21] J. Effah, P. K. Senyo, and S. Opoku-Anokye, "Business intelligence architecture informed by organisational semiotics," in International Conference on Informatics and Semiotics in Organisations, pp. 268–277, Springer, 2018.

[22] D. H. Fudholi and L. Mutawalli, "An ontology model for clinical pathway audit," in 2018 4th International Conference on Science and Technology (ICST), pp. 1–6, IEEE, 2018.

[23] H.-Q. Wang, T.-S. Zhou, Y.-F. Zhang, L. Chen, and J.-S. Li, "Research and development of semantics-based sharable clinical pathway systems," Journal of medical systems, vol. 39, no. 7, p. 73, 2015.

[24] J. Liu, Z. Huang, X. Lu, and H. Duan, "An ontology-based real-time monitoring approach to clinical pathway," in 2014 7th International Conference on Biomedical Engineering and Informatics, pp. 756–761, IEEE, 2014.

[25] S. R. Abidi and S. S. R. Abidi, "An ontological modeling approach to align institution-specific clinical pathways: Towards inter-institution care standardization," in 2012 25th IEEE International Symposium on Computer-Based Medical Systems (CBMS), pp. 1–4, IEEE, 2012.

[26] S. R. Abidi, S. S. R. Abidi, L. Butler, and S. Hussain, "Operationalizing prostate cancer clinical pathways: An ontological model to computerize, merge and execute institution-specific clinical pathways," in Workshop on Knowledge Management for Health Care Procedures, pp. 1–12, Springer, 2008.

[27] A. Daniyal, S. R. Abidi, and S. S. R. Abidi, "Computerizing clinical pathways: ontology-based modeling and execution.," in MIE, pp. 643–647, 2009.

[28] Z. Hu, J.-S. Li, T.-S. Zhou, H.-Y. Yu, M. Suzuki, and K. Araki, "Ontology-based clinical pathways with semantic rules," Journal of medical systems, vol. 36, no. 4, pp. 2203–2212, 2012.

[29] D. A. Alexandrou, K. V. Pardalis, T. D. Bouras, P. Karakitsos, and G. N. Mentzas, "Sempath ontology: Modeling multidisciplinary treatment schemes utilizing semantics," IEEE Transactions on Information Technology in Biomedicine, vol. 16, no. 2, pp. 235–240, 2011.

[30] Y. Ye, Z. Jiang, X. Diao, D. Yang, and G. Du, "An ontology-based hierarchical semantic modeling approach to clinical pathway workflows," Computers in biology and medicine, vol. 39, no. 8, pp. 722–732, 2009.

[31] Z. Hu, J.-S. Li, H.-y. Yu, X.-g. Zhang, M. Suzuki, and K. Araki, "Modeling of clinical pathways based on ontology," in 2009 IEEE International Symposium on IT in Medicine & Education, vol. 1, pp. 1170–1174, IEEE, 2009.

[32] A. Alahmar, Matteo Crupi, and Rachid Benlamri, Ontological Framework for Standardizing and Digitizing Clinical Pathways in Healthcare Information Systems, Computer Methods and Programs in Biomedicine, Elsevier, Vol. 196, 2020, pp. 1-18. DOI: 10.1016/j.cmpb.2020.105559.

[33] Ayman Alahmar and Rachid Benlamri, SNOMED CT-Based Standardized e-Clinical Pathways for Enabling Big Data Analytics in Healthcare, IEEE Access, Vol. 8, 2020, pp. 92765-92775. DOI: 10.1109/ACCESS.2020.2994286.

[34] Ayman Alahmar and Rachid Benlamri, Optimizing Hospital Resources using Big Data Analytics with Standardized e-Clinical Pathways, Proc. of 6th IEEE Int. Conf. on Cloud

and Big Data Computing (CBCCom), Calgary, Canada, August 17-24, 2020, pp. 650-657. DOI: 10.1109/DASC-PICom-CBDCom-CyberSciTech49142.2020.00112

[35] J. Ma, R. Zhang, X. Zhu, R. Cao, Process ontology technology in modeling clinical pathway information system, International Journal of Computers and Applications, 2020 Aug 17;42(6):550-7.

[36] S. H. Hoelscher and S. McBride, "Digitizing infectious disease clinical guidelines for improved clinician satisfaction," CIN: Computers, Informatics, Nursing, 2020.

[37] P. B. Smulowitz, Y. Dizitzer, S. Tadiri, L. Thibodeau, L. Jagminas, and V. Novack, "Impact of implementation of the heart pathway using an electronic clinical decision support tool in a community hospital setting," The American journal of emergency medicine, vol. 36, no. 3, pp. 408–413, 2018.

[38] J. Gibbs, L. J. Sutcliffe, V. Gkatzidou, K. Hone, R. E. Ashcroft, E. M. Harding-Esch, C. M. Lowndes, S. T. Sadiq, P. Sonnenberg, and C. S. Estcourt, "The eclinical care pathway framework: a novel structure for creation of online complex clinical care pathways and its application in the management of sexually transmitted infections," BMC medical informatics and decision making, vol. 16, no. 1, p. 98, 2016.

[39] R. Blaser, M. Schnabel, C. Biber, M. B¨aumlein, O. Heger, M. Beyer, E. Opitz, R. Lenz, and K. A. Kuhn, "Improving pathway compliance and clinician perfor- mance by using information technology," International journal of medical informat- ics, vol. 76, no. 2-3, pp. 151–156, 2007.

[40] K. Bernstein and U. Andersen, "Managing care pathways combining snomed ct, archetypes and an electronic guideline system.," Studies in health technology and informatics, vol. 136, p. 353, 2008.

[41] I. L. Katzan, Y. Fan, M. Speck, J. Morton, L. Fromwiller, J. Urchek, K. Uchino, S. D. Griffith, and M. Modic, "Electronic stroke carepath: integrated approach to stroke care," Circulation: Cardiovascular Quality and Outcomes, vol. 8, no. 6 suppl 3, pp. S179–S189, 2015.

[42] "Epic Systems Corporation." https://www.epic.com. Accessed: 2021-01-30.

[43] A. Alahmar, M. Almousa, and R. Benlamri, Automated Clinical Pathway Standardization using SNOMED CT Based Semantic Relatedness, unpublished (submitted for publication).